\documentclass{article} 
\usepackage[preprint]{colm2026_conference}

\usepackage{microtype}
\usepackage{hyperref}
\usepackage{url}
\usepackage{booktabs}
\usepackage{graphicx}
\usepackage{subcaption}
\usepackage{wrapfig}
\usepackage{amsmath,amssymb}
\usepackage{xcolor}
\usepackage{xspace}
\usepackage{multirow}
\usepackage{algorithm}
\usepackage{algorithmic}
\usepackage{pifont}
\usepackage{lineno}

\definecolor{darkblue}{rgb}{0, 0, 0.5}
\definecolor{burntorange}{rgb}{0.78, 0.36, 0.12}
\hypersetup{colorlinks=true, citecolor=darkblue, linkcolor=darkblue, urlcolor=darkblue}

\usepackage[utf8]{inputenc}
\usepackage[T1]{fontenc}

\newcommand{\ours}{MemRouter\xspace}

\title{MemRouter: Memory-as-Embedding Routing for Long-Term Conversational Agents}

\author{%
\\[-1em]
\hspace*{0.01\textwidth}
\begin{tabular}{@{}p{0.48\textwidth}@{\hspace{0.02\textwidth}}p{0.48\textwidth}@{}}
{\normalfont\bfseries Tianyu Hu} &
{\normalfont\bfseries Weikai Lin} \\
{\normalfont\mdseries University of Central Florida} &
{\normalfont\mdseries University of Rochester} \\
{\normalfont\mdseries\ttfamily ti721494@ucf.edu} &
{\normalfont\mdseries\ttfamily wlin33@ur.rochester.edu}
\\[2.0em]
{\normalfont\bfseries Weizhi Zhang} &
{\normalfont\bfseries Jing Ma} \\
{\normalfont\mdseries University of Illinois at Chicago} &
{\normalfont\mdseries Case Western Reserve University} \\
{\normalfont\mdseries\ttfamily wzhan42@uic.edu} &
{\normalfont\mdseries\ttfamily jing.ma5@case.edu}
\\[2.0em]
{\normalfont\bfseries Song Wang\thanks{Corresponding author.}} & \\
{\normalfont\mdseries University of Central Florida} & \\
{\normalfont\mdseries\ttfamily song.wang@ucf.edu} & 
\end{tabular}%
}

\begin{document}

\ifcolmsubmission
\linenumbers
\fi

\maketitle

\begin{abstract}
Long-term conversational agents must decide which turns to store in external memory, yet recent systems rely on autoregressive LLM generation at every turn to make that decision. We present MemRouter, a write-side memory router that decouples memory admission from the downstream answer backbone and replaces per-turn memory-management decoding with an embedding-based routing policy. MemRouter encodes each turn together with recent context, projects the resulting embeddings through a frozen LLM backbone, and predicts whether the turn should be stored using lightweight classification heads while training only ${\sim}$12M parameters. Under a controlled matched-harness comparison on LoCoMo, where the retrieval pipeline, answer prompts, and QA backbone (Qwen2.5-7B) are held identical, MemRouter outperforms an LLM-based memory manager on every question category (overall F1 52.0 vs.\ 45.6, non-overlapping 95\% CIs) while reducing memory-management p50 latency from 970\,ms to 58\,ms. Descriptive factorial averaging further shows that learned admission improves mean F1 by +10.3 over random storage, category-specific prompting adds +5.2 over a generic prompt, and retrieval contributes +0.7. These results suggest that write-side memory admission can be learned by a small supervised router, while answer generation remains a separate downstream component in long-horizon conversational QA. The code is available at \href{https://github.com/SongW-SW/MemRouter}{https://github.com/SongW-SW/MemRouter}.
\end{abstract}

\section{Introduction}

Long-term conversational memory is becoming a critical capability for LLM agents deployed in real-world applications~\citep{zhang2024memorysurvey, liu2025memorysurvey, pink2025episodic}. In multi-session settings, agents must selectively store personal facts, preferences, plans, and emotional states in an external memory bank and retrieve them accurately later~\citep{maharana2024locomo, zhong2024memorybank, packer2023memgpt, xu2025amem, nan2025nemori, chhikara2025mem0}. Selective admission matters not only for answer quality, but also for controlling memory growth over time and reducing unnecessary retention of personal information. As conversations grow to hundreds of turns and tens of thousands of tokens, the fundamental question becomes: \emph{how should an agent decide what to remember, and at what cost?}

Recent work uses the LLM itself as the memory manager~\citep{yan2025memoryr1, yu2026agemem, zhang2026amac, packer2023memgpt, xu2025amem}. Memory-R1~\citep{yan2025memoryr1} trains an RL-based memory manager over ADD/UPDATE/DELETE/NOOP operations together with an answer agent, achieving strong results with limited QA supervision. AgeMem~\citep{yu2026agemem} exposes six memory operations as tool calls and learns a unified policy via three-stage progressive RL. A-MAC~\citep{zhang2026amac} decomposes memory admission into interpretable factors including utility, confidence, and novelty. MemGPT~\citep{packer2023memgpt} employs an OS-inspired paging mechanism where the LLM manages its own context through function calls. Personalized dialogue systems such as Hello Again!~\citep{li2025helloagain}, RMM~\citep{tan2025rmm}, and MemInsight~\citep{salama2025meminsight} highlight the importance of long-term memory management in realistic interactions.

While effective, these approaches share two practical limitations. \textbf{First, per-turn autoregressive cost}: every conversational turn requires one or more full LLM generation calls to make the admission decision. Memory-R1, for instance, uses two generation calls per turn, one for fact extraction and one for the memory manager, resulting in approximately 1,200 generation calls for a single 600-turn conversation, \emph{before any questions are answered}. For a system processing hundreds of conversations daily, this write-side overhead can dominate the total serving budget. The underlying question is whether a full autoregressive pass is truly necessary to decide if a turn is worth storing, or whether this decision can be made more cheaply at the embedding level. \textbf{Second, backbone coupling}: because write-side policies are often trained jointly with, or against rewards defined by, the downstream answer model, the admission policy can become entangled with a specific answer backbone. Upgrading to a newer or larger QA model may then require retraining or re-tuning the memory management policy. An ideal write-side router would operate independently of the downstream answer agent, so that it can be trained once and reused across different QA backends without modification.

More broadly, prior agent-memory and external-memory architectures suggest that persistence can emerge from retrieved memory streams, explicit long-term memory banks, or compressed dialogue summaries~\citep{park2023generative, wang2023longmem, wang2023recursive}. What remains less explored is whether the write-side admission decision itself can be isolated and learned without invoking autoregressive generation at every turn.

We address both limitations with \textbf{\ours}, an embedding-based memory router that makes per-turn storage decisions entirely in embedding space. Instead of generating memory actions in text, the router encodes recent conversational context into a short sequence of chunks, maps those chunk embeddings through a frozen LLM backbone, and predicts whether the current turn should be stored using lightweight classification heads. This removes per-turn autoregressive decoding from the write path while keeping the admission policy separate from the downstream answer agent, so the same router can be reused across QA backbones. As we show later, this design yields both higher matched-harness accuracy and substantially lower write-side latency than an LLM-based manager.

Our contributions are threefold: an embedding-based write-side router that replaces per-turn memory-management decoding with a forward-only classifier built from a frozen backbone plus ${\sim}$12M trainable parameters; a supervised training pipeline using teacher-generated turn-level labels without reinforcement learning while preserving clean conversation-level data splits; and controlled experiments showing that, under identical retrieval and prompting conditions, \ours's embedding-based admission outperforms an LLM-based manager and stronger matched-budget baselines.

\section{Related Work}

\paragraph{Memory Architectures for LLM Agents.}
The need for external memory in LLM agents is widely recognized, because context windows, although growing, remain insufficient for long-horizon tasks. MemGPT~\citep{packer2023memgpt} pioneered the analogy between OS virtual memory and LLM memory management, introducing a paging mechanism between main context and external storage. A-MEM~\citep{xu2025amem} organizes memories through Zettelkasten-inspired indexing, MemoryBank~\citep{zhong2024memorybank} models temporal decay and reinforcement, and Mem0~\citep{chhikara2025mem0} provides a modular stack that combines graph-based and embedding-based retrieval. Generative Agents~\citep{park2023generative} maintains a retrieved memory stream with reflection and planning, while LONGMEM~\citep{wang2023longmem} augments a language model with a long-term memory bank. Compression-oriented dialogue memory methods such as recursive summarization~\citep{wang2023recursive} and compressive memory for long conversations~\citep{chen2024comedy} also target long-horizon interaction by condensing prior dialogue. Personalized dialogue systems such as Hello Again!~\citep{li2025helloagain}, RMM~\citep{tan2025rmm}, and MemInsight~\citep{salama2025meminsight} highlight the importance of long-term memory in realistic interactions. These systems advance memory \emph{organization and retrieval}, but they typically depend on heuristic rules or the LLM itself for the upstream decision of \emph{what to store}, the bottleneck our work targets.

\paragraph{RL-Trained Memory Management.}
A recent line of work applies reinforcement learning to train LLMs for adaptive memory management. Memory-R1~\citep{yan2025memoryr1} is the closest comparison to our setting: it trains both a Memory Manager, which predicts ADD/UPDATE/DELETE/NOOP operations, and an Answer Agent, which distills retrieved memories before answering, using PPO and GRPO with exact-match reward. Memory-R1 achieves strong results with very limited QA supervision, but both agents are full 7B LLMs and the Memory Manager still performs per-turn generation. AgeMem~\citep{yu2026agemem} exposes six memory operations as tool calls and trains a unified policy through progressive RL, while A-MAC~\citep{zhang2026amac} decomposes admission into interpretable factors such as utility, novelty, and confidence and explicitly studies the latency tradeoffs of this more structured decision process. Related systems such as Mem-$\alpha$~\citep{wang2025memalpha}, HiAgent~\citep{hu2024hiagent}, and ACON~\citep{kang2025acon} also treat memory construction or compression as a learned policy. Our work departs from this line by asking whether the LLM needs to participate in per-turn admission decisions at all.

\paragraph{Routing and Gating Mechanisms for Memory.}
Router-based architectures have proven effective across many domains, from Mixture-of-Experts~\citep{shazeer2017outrageously} for efficient LLM inference to specialized routers in retrieval-augmented generation. In the memory domain, RCR-Router~\citep{liu2025rcrrouter} routes memory subsets to different agents based on role and task stage, MemR3~\citep{du2025memr3} routes among retrieve, reflect, and answer actions, G-MemLLM~\citep{xu2026gmemllm} introduces differentiable gates for latent memory updates, and FluxMem~\citep{fluxmem2026} selects among multiple memory structures through a learned gate. These works show the value of routing in memory systems, but they primarily route \emph{retrieval} or latent update behavior. Our work instead applies routing to the \emph{write-side} decision of whether a conversational turn should enter long-term memory, and does so at the embedding level rather than through autoregressive generation.


\begin{figure*}[ht]
\centering
\includegraphics[width=\textwidth]{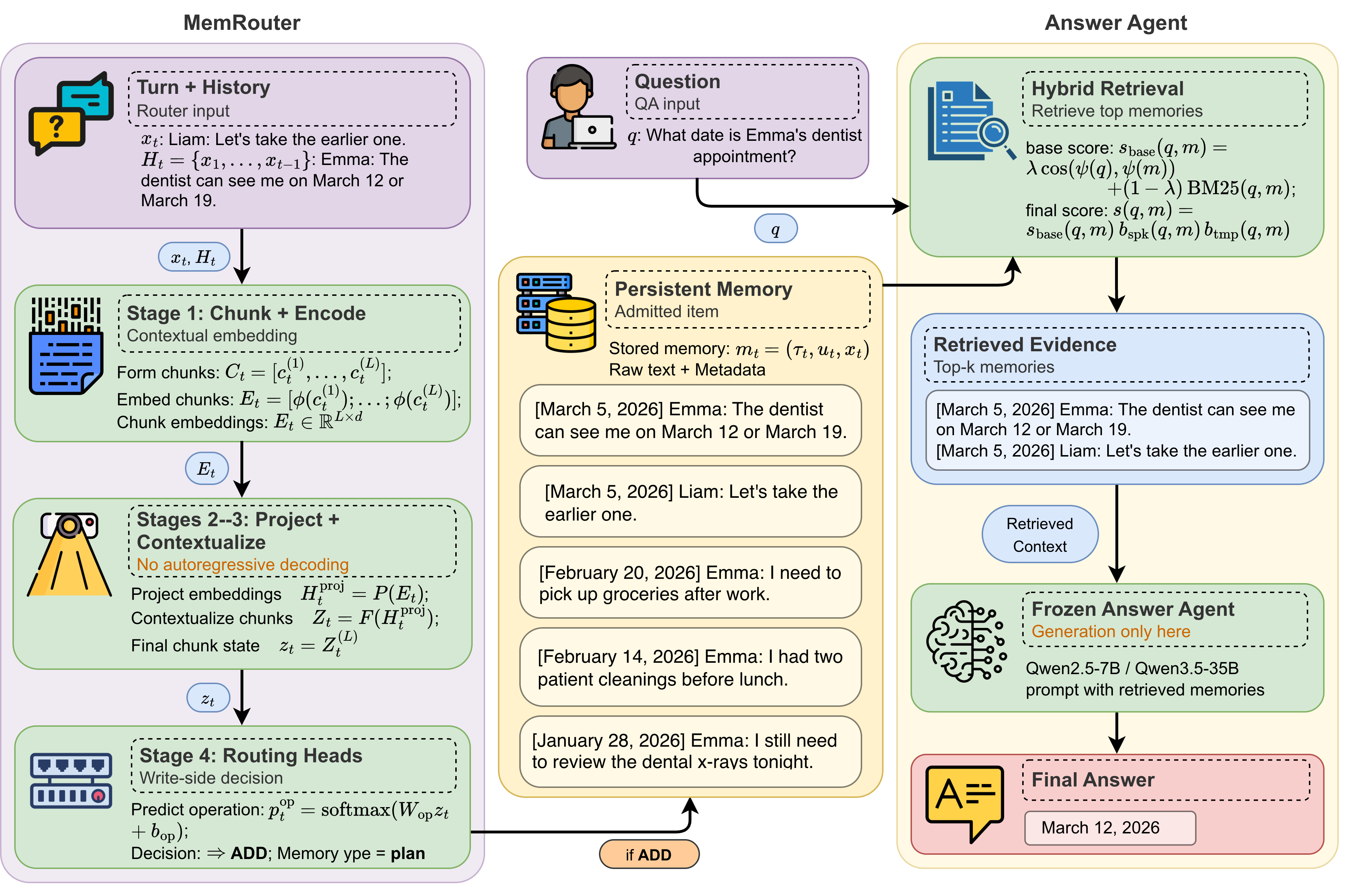}
\caption{\textbf{MemRouter system overview.} \textbf{Left: Write path.} Given turn $x_t$ and history $H_t$, the router forms contextual chunks, encodes them into dense embeddings, projects them into the frozen backbone space, contextualizes them, and predicts whether the turn should be written to memory. If the decision is ADD, the admitted item is stored in persistent memory. \textbf{Center: Persistent memory.} Stored entries retain verbatim turn text together with timestamp and speaker metadata. \textbf{Right: Read / QA path.} At question time, hybrid retrieval ranks the stored memories and returns the top evidence to a frozen answer agent, which performs the only text-generation step in the pipeline.}
\label{fig:memrouter_system}
\end{figure*}

\section{Method}

\subsection{Overview}

The central design principle of \ours is to separate the \emph{decision} of what to remember from the \emph{capability} of answering questions. Memory management decisions, which are made for every conversational turn, should avoid per-turn autoregressive decoding overhead. Question answering can still use full LLM generation. This asymmetry motivates our three-component architecture, summarized in Figure~\ref{fig:memrouter_system}, which separates the per-turn write path from the question-time read and answer path:
\textbf{(1) Memory Router} processes every conversation turn to decide whether it should be stored. Since this component is invoked hundreds of times per conversation, it should avoid per-turn autoregressive decoding. We implement it as an embedding-based classifier built on frozen components.
\textbf{(2) Memory Store} maintains the stored entries with dense embeddings for retrieval. It combines semantic matching with keyword search to support diverse question types (factual, temporal, entity-specific).
\textbf{(3) Answer Agent} is invoked when the user poses a question. It retrieves relevant memories and generates an answer using a frozen LLM.

\subsection{Memory Router Architecture}

The router must process each turn efficiently while still capturing enough semantic content to make informed storage decisions. Our design leverages a frozen LLM backbone's representational power while avoiding per-turn text-generation overhead. Given a conversation turn $x_t$ with history $H_t = \{x_1, \ldots, x_{t-1}\}$, the router proceeds in three stages:

\paragraph{Stage 1: Contextual Embedding and Projection.}
To make an informed storage decision, the router must understand the current turn \emph{in context}: a turn saying ``Let's take the earlier one'' is unimportant on its own but may confirm a concrete plan when considered with preceding turns. We first partition the recent dialogue context into a sequence of fixed-size chunks
\begin{equation}
    C_t = [c_t^{(1)}, \ldots, c_t^{(L)}],
\end{equation}
where each chunk contains up to 5 turns and $L \leq 13$ (the current chunk plus up to 12 history chunks). We then encode each chunk with a sentence transformer $\phi$ (BGE-large-en-v1.5~\citep{xiao2023cpack}, 335M parameters, 1024-dimensional output):
\begin{equation}
    E_t = [\phi(c_t^{(1)}); \ldots; \phi(c_t^{(L)})] \in \mathbb{R}^{L \times d}.
\end{equation}
Here $d = 1024$, and the chunking strategy balances context breadth with computational efficiency: each chunk captures a coherent conversational segment, while the 12-chunk history window covers the most recent 60 turns. These sentence embeddings are then mapped to the backbone hidden dimension ($d' = 3584$ for Qwen2.5-7B) with a two-layer projection that applies LayerNorm and GELU:
\begin{equation}
    H_t = P(E_t) \in \mathbb{R}^{L \times d'},
\end{equation}
where $P: \mathbb{R}^{d} \rightarrow \mathbb{R}^{d'}$ is a two-layer MLP with LayerNorm and GELU applied row-wise to each chunk embedding. This projection constitutes ${\sim}$8M trainable parameters and is the primary learned component for aligning the embedding space with the backbone's representation space during router training.

\paragraph{Stage 2: Backbone Contextualization.}
The projected chunk embeddings are fed directly into the frozen Qwen2.5-7B transformer body~\citep{qwen2024qwen25} via its continuous input interface instead of the standard token-embedding lookup. This design does \emph{not} eliminate neural inference: the router still performs a frozen-backbone forward pass for each turn. However, it avoids tokenization-driven autoregressive generation, which is the dominant per-turn cost in Memory-R1-style memory management. In the evaluated setting, the router operates on an unpadded variable-length chunk sequence ordered chronologically from retained history to the current turn, so no explicit attention mask is required at inference time. We do not introduce additional input scaling beyond the projection module itself, which already applies LayerNorm and GELU before the projected chunk sequence is cast to the backbone computation dtype. The backbone's 28 transformer layers then contextualize the chunk sequence in a single forward pass, enabling the model to capture dependencies between the current turn and its conversational context:
\begin{equation}
    Z_t = F(H_t) \in \mathbb{R}^{L \times d'}, \qquad z_t = Z_t^{(L)},
\end{equation}
where $F$ denotes the frozen transformer body and $z_t$ is the final chunk representation corresponding to the current turn. Critically, all backbone parameters are \emph{frozen}. We therefore leverage the pre-trained transformer's representational capacity without modifying it, and the backbone can be shared with the answer agent without interference.

\paragraph{Stage 3: Operation Classification.}
The pooled representation is passed through lightweight classification heads to produce the routing decision:
\begin{equation}
    p_t^{\mathrm{op}} = \operatorname{softmax}(W_{\mathrm{op}} z_t + b_{\mathrm{op}}), \qquad
    p_t^{\mathrm{type}} = \operatorname{softmax}(W_{\mathrm{type}} z_t + b_{\mathrm{type}}).
\end{equation}
The operation head predicts a distribution over $\{\text{ADD}, \text{NOOP}\}$, while the content-type head predicts one of $\{\text{key\_facts}, \text{emotional}, \text{preference}, \text{plan}, \text{routine}\}$ and is only used when the target operation is ADD. Together, these heads add ${\sim}$4M parameters. The total trainable parameter count is ${\sim}$12M (projection + heads), which is 0.17\% of the full 7B model.

\subsection{Router Training}

Training requires turn-level supervision indicating whether a turn should be stored and, if stored, which content type it expresses. We obtain these labels from a teacher model (Qwen3.5-35B-A3B~\citep{qwenteam2026qwen35}) applied to dialogue turns from LoCoMo, LongMemEval~\citep{wu2024longmemeval}, and MSC. For LoCoMo, teacher labeling follows the same conversation-level split as QA evaluation: the first conversation is used for training reference, the second for validation and model selection, and the remaining eight are held out for testing only. No turns from LoCoMo test conversations are used for teacher labeling, router training, threshold tuning, prompt tuning, or model selection.

We train only the projection layers and classification heads, while keeping the sentence encoder and LLM backbone frozen. The objective combines weighted cross-entropy for the operation head with standard cross-entropy for the content-type head, applied only when the target operation is ADD:
\begin{equation}
    \mathcal{L} = \frac{1}{N} \sum_{t=1}^{N} \left[
    \mathrm{CE}_{w}\!\left(y_t^{\mathrm{op}}, p_t^{\mathrm{op}}\right)
    + \mathbf{1}[y_t^{\mathrm{op}}{=}\mathrm{ADD}] \cdot \mathrm{CE}\!\left(y_t^{\mathrm{type}}, p_t^{\mathrm{type}}\right)
    \right].
\end{equation}

In practice, turn embeddings are pre-computed and cached before optimization, so training operates on fixed embeddings rather than repeatedly re-encoding text. We also explored GRPO-style RL fine-tuning, but found supervised training more stable and effective for this lightweight router; quantitative comparisons appear in Section~\ref{sec:ablation}.

\subsection{Memory Store and Retrieval}

When the router decides ADD for a turn, we store the tuple
\begin{equation}
    m_t = (\tau_t, u_t, x_t),
\end{equation}
where $\tau_t$ is the session timestamp, $u_t$ is the speaker identity, and $x_t$ is the verbatim turn text. At indexing time, this tuple is serialized as \texttt{[session\_datetime] speaker: turn\_text}, making the timestamp part of the searchable content. We found that storing raw conversation text outperforms extracting structured facts via an LLM in this setting, because verbatim storage preserves wording that matches gold answers during token-level F1 evaluation.

At question-answering time, we retrieve the top-60 most relevant memories using a hybrid scoring function that combines dense semantic matching with BM25, a sparse retrieval function based on weighted keyword overlap:
\begin{equation}
    s_{\mathrm{base}}(q, m) = \lambda \cdot \cos(\psi(q), \psi(m)) + (1-\lambda) \cdot \mathrm{BM25}(q, m), \qquad \lambda = 0.7,
\end{equation}
\begin{equation}
    s(q, m) = s_{\mathrm{base}}(q, m) \cdot b_{\mathrm{spk}}(q, m) \cdot b_{\mathrm{tmp}}(q, m).
\end{equation}
Here $\psi$ denotes the retrieval encoder instantiated from the same BGE model family. Dense cosine similarity captures semantic relatedness (e.g., ``dental checkup'' matching ``dentist appointment''), while BM25 captures exact lexical matches that dense retrieval often misses (e.g., specific dates ``March 12, 2026'' or proper names). We normalize the dense and sparse scores separately and combine them with $\lambda = 0.7$.

We further apply lightweight query-dependent adjustments. If a question explicitly mentions a speaker, memories from that speaker receive an additional boost, with a stronger boost for inferential questions. If a question contains temporal cues, memories with timestamp metadata are also upweighted. Finally, a session diversity constraint limits over-retrieval from any single session, improving coverage across the full conversation.

\subsection{Answer Generation}

The answer agent is a frozen LLM (Qwen2.5-7B-Instruct~\citep{qwen2024qwen25} or Qwen3.5-35B-A3B~\citep{qwenteam2026qwen35}) that receives the top-60 retrieved memories and generates an answer. This is the \emph{only} component that performs LLM text generation, and it is invoked only once per question, not per turn. The agent is never fine-tuned; all its behavior is controlled through prompting.

We present retrieved memories grouped by speaker with timestamps, following the format used by Memory-R1~\citep{yan2025memoryr1}. This grouping helps the model associate information with specific speakers and attend to temporal ordering within each speaker's history. We use category-specific prompts based on the question type:

For \textbf{single-hop, temporal, and open-domain} questions, the prompt instructs the model to ``answer in 5--6 words'' and ``pay special attention to timestamps.'' The word limit is critical: it forces concise answers with less extraneous text, which achieves higher precision in token-level F1 evaluation. For temporal questions specifically, the prompt adds ``answer with specific dates (e.g., March 12, 2026), NOT relative terms like `next Thursday','' because we observed that models tend to copy relative temporal expressions from the conversation text rather than resolving them to absolute calendar dates.

For \textbf{multi-hop} questions, the prompt removes the word limit and instead instructs ``list all relevant items separated by commas,'' since multi-hop answers are typically enumerations (e.g., ``pottery, hiking, photography, volunteering'') that require more tokens. This category-specific prompting substantially improves multi-hop performance: F1 increases from 26.1 to 52.4 when switching from the default prompt to the multi-hop prompt.

\section{Experiments}

\subsection{Setup}

We evaluate on LoCoMo~\citep{maharana2024locomo}, following Memory-R1's 1:1:8 split and excluding adversarial questions. For MemRouter, LoCoMo supervision follows the same conversation-level split: conversation 1 is used for training, conversation 2 for validation, and conversations 3--10 are held out for test only. No LoCoMo test conversations are used in teacher labeling, router training, threshold tuning, prompt tuning, or model selection. Our primary baseline is Memory-R1~\citep{yan2025memoryr1}; we compare against their reported Qwen2.5-7B-Instruct results.

We adopt the official LoCoMo token-level F1 metric. Detailed benchmark statistics, split counts, metric definitions, baseline descriptions, and training hyperparameters are deferred to Appendix~\ref{app:datasets} and Appendix~\ref{app:impl}.

\subsection{Main Results}

Table~\ref{tab:main} compares the end-to-end \ours system against published baselines on the LoCoMo test set. All baseline numbers (LoCoMo RAG through Memory-R1-GRPO) are taken from~\citet{yan2025memoryr1} under their Qwen2.5-7B-Instruct evaluation, so the first \ours row is the like-for-like 7B comparison. The final row is shown separately as a backbone-transfer result that swaps only the answer agent to Qwen3.5-35B-A3B, with no router retraining. Note that the baseline systems each use their own retrieval and prompting pipelines, so these numbers reflect full system-level performance rather than a controlled comparison of admission policies alone (see Section~\ref{sec:controlled_comparison} for the latter).

\begin{table}[t]
\centering
\small
\begin{tabular}{@{}lccccc@{}}
\toprule
\textbf{Method} & \textbf{Overall} & \textbf{Single} & \textbf{Multi} & \textbf{Temp.} & \textbf{Open} \\
\midrule
\multicolumn{6}{@{}l}{\textit{Baselines from~\citet{yan2025memoryr1}, Qwen2.5-7B-Instruct:}} \\
LoCoMo (RAG) & 9.0 & 9.6 & 11.8 & 8.4 & 8.7 \\
A-Mem~\citep{xu2025amem} & 26.1 & 19.0 & 14.7 & 23.7 & 30.6 \\
Mem0~\citep{chhikara2025mem0} & 30.6 & 25.0 & 20.3 & 33.2 & 32.7 \\
MemoryOS~\citep{kang2026memoryos} & 34.6 & 29.6 & 21.0 & 26.3 & 40.9 \\
Memory-R1-GRPO~\citep{yan2025memoryr1} & 43.1 & 33.6 & 23.6 & \textbf{47.8} & \textbf{46.9} \\
\midrule
\ours (Qwen2.5-7B-Instruct) & \textbf{52.0} & \textbf{57.5} & \textbf{52.4} & 44.0 & 27.1 \\
\cmidrule(lr){1-6}
\multicolumn{6}{@{}l}{\textit{Backbone transfer (same router)}} \\
\ours (Qwen3.5-35B-A3B answer agent) & 55.5 & 60.9 & 56.1 & 49.7 & 23.8 \\
\bottomrule
\end{tabular}
\caption{End-to-end results on LoCoMo test set (F1, excluding adversarial). Baseline numbers from~\citet{yan2025memoryr1} use Qwen2.5-7B-Instruct. The first \ours row is the like-for-like 7B comparison; the final row is reported separately as a backbone-transfer result that swaps only the answer agent to Qwen3.5-35B-A3B, with no router retraining. Each system uses its own retrieval and prompting pipeline.}
\label{tab:main}
\end{table}

In the like-for-like 7B comparison, \ours achieves the highest single-hop (57.5) and multi-hop (52.4) F1 among all methods. Open-domain F1 (27.1 with 7B) trails Memory-R1-GRPO (46.9), which we attribute to the frozen answer agent's limited inference capability rather than a routing failure. The separate 35B transfer row shows that swapping only the answer agent to Qwen3.5-35B-A3B (4-bit, ${\sim}$18GB) raises temporal F1 to 49.7 and overall F1 to 55.5. \emph{The router weights are not retrained}, demonstrating backbone-agnostic transfer.

These end-to-end numbers combine write-side admission with each system's retrieval and prompting pipeline. Section~\ref{sec:controlled_comparison} therefore reports a matched-harness comparison that fixes the model, retrieval, and prompts and isolates the admission policy itself.

\paragraph{Backbone Transfer to LLaMA-3.1-8B.}

To validate that \ours generalizes across LLM backbones, we retrain only the router's lightweight heads (${\sim}$12M parameters) with a frozen LLaMA-3.1-8B-Instruct backbone and use the same model as the answer agent. Table~\ref{tab:llama_transfer} reports the results alongside a store-all baseline under the same backbone.

\begin{table}[t]
\centering
\small
\begin{tabular}{@{}lccccc@{}}
\toprule
\textbf{Method} & \textbf{Overall} & \textbf{Single} & \textbf{Multi} & \textbf{Temp.} & \textbf{Open} \\
\midrule
Store-all (LLaMA-3.1-8B) & 50.6 & 60.4 & 38.3 & 42.3 & 26.2 \\
\ours (LLaMA-3.1-8B) & 49.1 & 57.0 & 36.2 & \textbf{46.3} & 24.9 \\
\midrule
\ours (Qwen2.5-7B) & 52.0 & 57.5 & 52.4 & 44.0 & 27.1 \\
\bottomrule
\end{tabular}
\caption{Backbone transfer to LLaMA-3.1-8B-Instruct. The router heads are retrained with the LLaMA backbone frozen; the same LLaMA model serves as the answer agent. F1 excluding adversarial. The Qwen row is reproduced from Table~\ref{tab:main} for reference.}
\label{tab:llama_transfer}
\end{table}

With LLaMA-3.1-8B, \ours achieves an overall F1 of 49.1, within 3 points of the Qwen2.5-7B result (52.0), despite using a completely different backbone family. Temporal F1 (46.3) actually exceeds the Qwen result (44.0), approaching Memory-R1's 47.8. The multi-hop gap (36.2 vs.\ 52.4) is partly attributable to the router's higher storage rate (${\sim}$70\% vs.\ ${\sim}$62\%), which introduces retrieval noise for multi-hop queries that require combining information across fewer, more relevant memories. These results confirm that the router architecture transfers across backbone families with only head retraining, consistent with our backbone-agnostic design goal.

\subsection{Controlled Comparison with an LLM Manager}
\label{sec:controlled_comparison}

\begin{figure*}[th]
\centering
\begin{subfigure}[t]{0.48\textwidth}
\centering
\includegraphics[width=\linewidth]{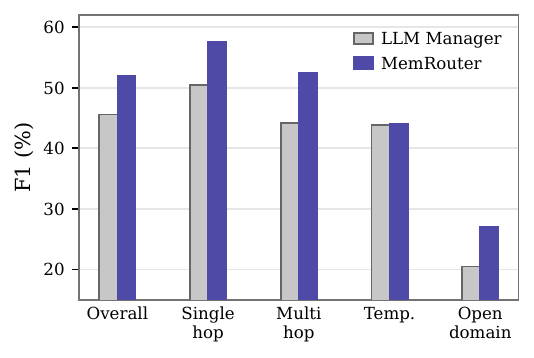}
\caption{Matched-harness F1 comparison.}
\label{fig:matched_subfig}
\end{subfigure}\hfill
\begin{subfigure}[t]{0.48\textwidth}
\centering
\includegraphics[width=\linewidth]{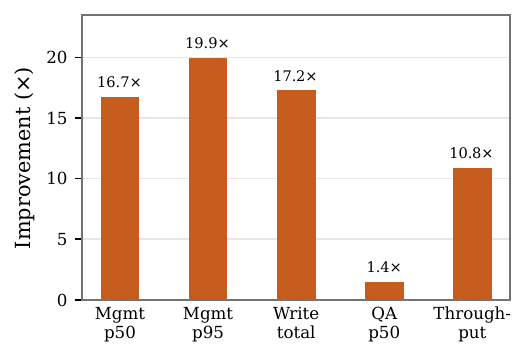}
\caption{Relative efficiency improvements.}
\label{fig:efficiency_subfig}
\end{subfigure}
\caption{Controlled comparison with an LLM-based memory manager, where the only difference is the write-side admission policy. Panel (a) reports matched-harness F1 excluding adversarial questions, with overall 95\% bootstrap CIs reported in the text and Appendix Table~\ref{tab:matched_appendix}. Panel (b) reports relative efficiency improvements, expressed as how many times \ours improves over the LLM manager on memory-management p50 latency, memory-management p95 latency, total write-path time, QA p50 latency, and end-to-end throughput; exact raw values are reported in Appendix Table~\ref{tab:efficiency_appendix}.}
\label{fig:controlled_comparison}
\end{figure*}

To isolate the effect of the write-side admission policy, we compare \ours against a local Memory-R1-style LLM-based memory manager baseline under a matched harness: both methods use the \emph{same} Qwen2.5-7B model, \emph{same} hybrid retrieval, and \emph{same} category-specific prompts, and differ only in how they decide whether to store each turn. Figure~\ref{fig:controlled_comparison}\subref{fig:matched_subfig} summarizes the matched-harness F1 comparison, while Figure~\ref{fig:controlled_comparison}\subref{fig:efficiency_subfig} shows the relative wall-clock efficiency comparison.

Under these controlled conditions, \ours outperforms the LLM-based manager on \emph{every} question category, with the largest gains on single-hop (+14\%) and multi-hop (+19\%). The overall 95\% bootstrap confidence intervals do not overlap ($[51, 53]$ vs.\ $[42, 49]$), indicating a reliable difference. The LLM manager also shows higher per-conversation variance (std\,=\,5.5 vs.\ 2.1), suggesting less consistent memory decisions across conversation styles.

\paragraph{Wall-Clock Efficiency.}
Using the same matched harness, we also measure wall-clock efficiency. Figure~\ref{fig:controlled_comparison} (right) plots the relative improvement factors for the main latency and throughput metrics, while Appendix Table~\ref{tab:efficiency_appendix} reports the full latency, throughput, and GPU-memory numbers.

The per-turn memory management latency is \textbf{17$\times$ lower} for \ours (58\,ms vs.\ 970\,ms at p50), because the router performs a single frozen forward pass over pre-computed embeddings rather than autoregressive text generation. This translates to a \textbf{17$\times$ reduction in total write-path wall-clock time} (293\,s vs.\ 5{,}028\,s) and an \textbf{11$\times$ improvement in end-to-end throughput} (2.58 vs.\ 0.24 QA/s). GPU peak memory is slightly higher for \ours (31.2 vs.\ 26.6\,GB) due to the additional BGE-large encoder and router projection heads.

\subsection{Ablation Studies}
\label{sec:ablation}
\label{sec:storage_policy}

\paragraph{Budget-Matched Storage Policies.}
Table~\ref{tab:storage_policies} reports the full storage-policy comparison under a matched 62\% storage budget.

\begin{table}[!t]
\centering
\small
\begin{tabular}{@{}lcccccc@{}}
\toprule
\textbf{Policy} & \textbf{Store\%} & \textbf{F1} & \textbf{Single} & \textbf{Multi} & \textbf{Temp.} & \textbf{Open} \\
\midrule
Store-all (upper bound) & 100\% & 53.7 & 60.5 & 53.7 & 43.8 & 25.6 \\
\midrule
\textbf{\ours} & \textbf{62\%} & \textbf{50.8} & \textbf{56.2} & \textbf{52.2} & 42.2 & \textbf{26.1} \\
MLP-only (no backbone) & 62\% & 50.0 & 54.7 & 51.7 & \textbf{43.3} & 24.9 \\
Keyword heuristic & 62\% & 47.2 & 50.5 & 52.1 & 41.8 & 21.3 \\
Random & 62\% & 42.8 & 45.8 & 44.9 & 38.5 & 24.1 \\
Recent-$k$ & 62\% & 38.6 & 42.9 & 42.6 & 27.8 & 21.8 \\
\bottomrule
\end{tabular}
\caption{Budget-matched storage policy comparison. All policies store ${\sim}$62\% of turns (except store-all) and use the same hybrid retrieval pipeline, speaker-grouped prompts, and Qwen2.5-7B answer agent. Only the write-side storage decision varies. For score-based policies (keyword, MLP, \ours), we rank turns by policy score and select the top 62\%. F1 excluding adversarial.}
\label{tab:storage_policies}
\end{table}

At that budget, \ours achieves the best overall F1 (50.8), ahead of MLP-only (50.0), keyword heuristic (47.2), random storage (42.8), and recency-based storage (38.6). Store-all remains the unconstrained upper bound at 53.7, indicating that learned admission matters most when storage is limited rather than when every turn can simply be retained.
This matched-budget view isolates storage selection quality from raw storage volume. It also shows that contextual routing retains an advantage over both heuristic and simpler learned policies when capacity is fixed.

\begin{wrapfigure}[15]{t}{0.4\textwidth}
\vspace{-1\baselineskip}
\centering
\includegraphics[width=\linewidth]{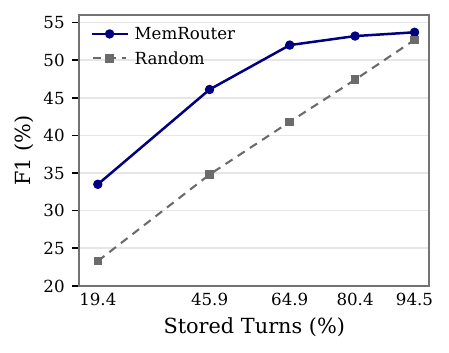}
\caption{Memory budget sweep: F1 (\%, excluding adversarial questions) versus storage ratio.}
\label{fig:budget_sweep}
\end{wrapfigure}

\paragraph{Memory Budget Curve.}

To characterize the accuracy--storage tradeoff, we sweep the router's ADD threshold from 0.1 to 0.9 and compare against random storage at each matched budget. This matched-budget comparison separates storage quality from the trivial advantage of simply retaining more turns under the same budget and storage volume. Figure~\ref{fig:budget_sweep} shows that \ours consistently outperforms random storage at every budget level in the sweep, with the gap \emph{widening} as the budget decreases: at 94.5\% storage the advantage is only 1.0 F1, but at 45.9\% it grows to 11.4. This demonstrates that the router's learned selectivity becomes increasingly valuable in resource-constrained settings where not all turns can be stored.

\begin{figure*}[h]
\centering
\includegraphics[width=0.9\textwidth]{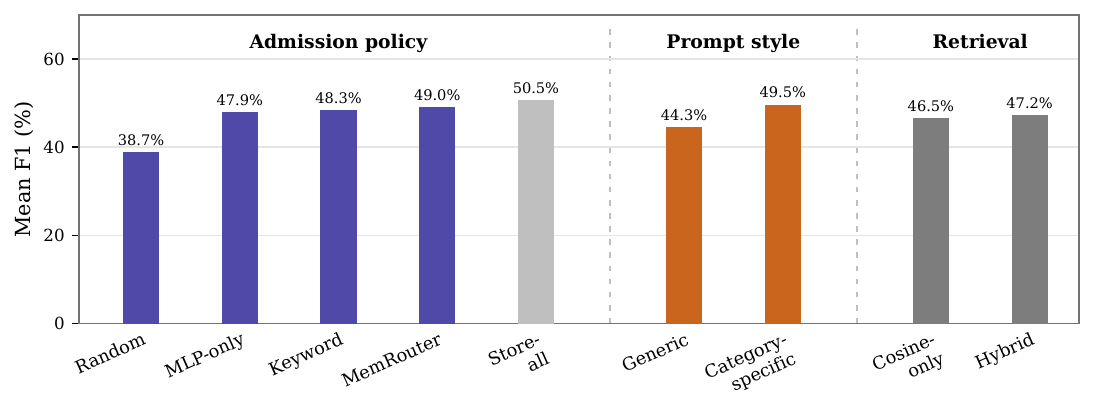}
\caption{Mean F1 (\%) for each admission policy, prompt style, and retrieval setting after averaging over the other two factors. Store-all is included as an unconstrained reference and is not budget-matched.}
\label{fig:factorial_bars}
\end{figure*}

\paragraph{Descriptive Factor Attribution.}
Figure~\ref{fig:factorial_bars} summarizes descriptive factor averages on a mean-F1 percentage scale. Admission policy spans the widest range, from 38.7 for random storage to 49.0 for MemRouter, with store-all reaching 50.5 as an unconstrained reference. Prompt style shifts mean F1 from 44.3 with a generic prompt to 49.5 with category-specific prompting, whereas retrieval changes only modestly from 46.5 for cosine-only search to 47.2 for hybrid retrieval. Exact marginal means are reported in Appendix Table~\ref{tab:factorial}.

Together with the end-to-end comparison against Memory-R1 (Table~\ref{tab:main}), the matched-harness comparison (Figure~\ref{fig:controlled_comparison}\subref{fig:matched_subfig}), the budget sweep (Figure~\ref{fig:budget_sweep}), and the descriptive factor chart (Figure~\ref{fig:factorial_bars}) collectively support the conclusion that learned write-side admission improves both accuracy and efficiency. Detailed retrieval/prompt ablations and additional supporting results are reported in Appendix~\ref{app:additional_results}.

\section{Conclusion}

We presented \ours, an embedding-based memory router that avoids per-turn autoregressive decoding in the memory-management loop while leaving answer generation to a separate downstream stage. Under a matched-harness LoCoMo evaluation with the same Qwen2.5-7B backbone, retrieval pipeline, and prompts, it outperforms an LLM-based memory manager on every question category while reducing memory-management p50 latency from 970\,ms to 58\,ms. These results show that write-side memory admission can be learned effectively as a small supervised router, enabling long-horizon conversational systems to build useful external memory without a full generation call on every turn.

\bibliography{ref}
\bibliographystyle{colm2026_conference}

\clearpage
\appendix

\section{Experimental Details}
This appendix first summarizes the dataset, evaluation protocol, and implementation settings used throughout the paper.

\subsection{Dataset and Evaluation Details}
\label{app:datasets}

\paragraph{Benchmark.}
We evaluate on LoCoMo~\citep{maharana2024locomo}, the standard benchmark for long-term conversational memory. The released benchmark consists of 10 multi-session conversations, each spanning approximately 35 sessions over several months, with an average of 588.2 turns and 16,618.1 tokens per conversation. Each conversation is annotated with approximately 200 QA pairs spanning five categories: single-hop (factual recall from one session), multi-hop (synthesizing across sessions), temporal (date/time reasoning), open-domain (inference from indirect evidence), and adversarial (questions about nonexistent information).

\paragraph{Data Split.}
Following the protocol established by Memory-R1~\citep{yan2025memoryr1}, we use a 1:1:8 train/validation/test split: the first conversation (152 QA pairs) for training reference, the second (81 QA pairs) for validation, and the remaining eight conversations (1,307 QA pairs after excluding adversarial) for testing. LoCoMo supervision follows this same split: conversation 1 is the only LoCoMo source used for training reference, conversation 2 is used for validation and model selection, and conversations 3--10 are held out for test only. We exclude adversarial questions from evaluation, consistent with Memory-R1, which notes that this subset ``lacks ground-truth answers'' suitable for F1 evaluation.

\paragraph{Metrics.}
We adopt the official LoCoMo evaluation protocol: token-level F1 with Porter stemming and answer normalization (lowercasing, removing articles including ``and'', removing punctuation and commas). Category-specific evaluation logic is applied: multi-hop questions use comma-split matching (each gold sub-answer is matched against the best prediction sub-answer); open-domain questions use the first answer before any semicolons; temporal and single-hop questions use standard F1.

\paragraph{Baselines.}
Our primary baseline is Memory-R1~\citep{yan2025memoryr1}, the current state-of-the-art on LoCoMo with RL-trained memory management. We use their reported numbers on Qwen2.5-7B-Instruct. We also report ablations that serve as internal baselines (dense-only retrieval, different embedding models, alternative training approaches).

\subsection{Implementation Details}
\label{app:impl}

\paragraph{Training Details.}
Router supervision is generated by a Qwen3.5-35B-A3B teacher model in 4-bit quantization over a unified label set of 28,415 turns drawn from LoCoMo, LongMemEval, and MSC. For LoCoMo, only conversation 1 contributes training-side supervision, conversation 2 is reserved for validation and model selection, and conversations 3--10 are excluded from teacher labeling and router training. We keep the sentence encoder and LLM backbone frozen and train only the projection and classification heads. All BGE embeddings are pre-computed once and cached before optimization. Unless otherwise noted, supervised router training uses 5 epochs, batch size 16, learning rate $10^{-3}$, and conversation-level validation for model selection. Training takes approximately 10 minutes on a single NVIDIA B200 GPU, while one-time label generation runs at about 4.8 turns per second and completes in approximately 98 minutes.

\section{Exact Numbers for Main-Paper Figures}
This section reports the exact numeric values underlying the main-paper comparison figures.

\paragraph{Matched-Harness Comparison.}
Table~\ref{tab:matched_appendix} reports the exact matched-harness numbers underlying Figure~\ref{fig:controlled_comparison}. Both methods use the same Qwen2.5-7B answer model, retrieval pipeline, and prompts, and differ only in the write-side admission policy.

\begin{table}[ht]
\centering
\small
\begin{tabular}{@{}lccccc@{}}
\toprule
\textbf{Method} & \textbf{Overall} & \textbf{Single} & \textbf{Multi} & \textbf{Temp.} & \textbf{Open} \\
\midrule
LLM Manager & 45.6 {\scriptsize [42,49]} & 50.4 & 44.2 & 43.8 & 20.5 \\
\ours & 52.0 {\scriptsize [51,53]} & 57.5 & 52.4 & 44.0 & 27.1 \\
\bottomrule
\end{tabular}
\caption{Exact values underlying Figure~\ref{fig:controlled_comparison}\subref{fig:matched_subfig}. Both methods use the same Qwen2.5-7B answer model, hybrid retrieval pipeline, and category-specific prompts; they differ only in the write-side admission policy. Overall 95\% bootstrap confidence intervals are shown in brackets.}
\label{tab:matched_appendix}
\end{table}

\paragraph{Wall-Clock Efficiency.}
Table~\ref{tab:efficiency_appendix} reports the exact wall-clock measurements underlying Figure~\ref{fig:controlled_comparison}. Both methods use the same Qwen2.5-7B model, retrieval pipeline, and prompts; they differ only in the write-side admission policy.

\begin{table}[ht]
\centering
\small
\begin{tabular}{@{}lcc@{}}
\toprule
\textbf{Metric} & \textbf{LLM Manager} & \textbf{\ours} \\
\midrule
Memory mgmt p50 latency (ms/turn) & 970 & \textbf{58} \\
Memory mgmt p95 latency (ms/turn) & 1{,}415 & \textbf{71} \\
Memory mgmt total time (s) & 5{,}028 & \textbf{293} \\
QA p50 latency (ms/question) & 198 & \textbf{141} \\
End-to-end throughput (QA/s) & 0.24 & \textbf{2.58} \\
GPU peak memory (GB) & 26.6 & 31.2 \\
\bottomrule
\end{tabular}
\caption{Exact values underlying Figure~\ref{fig:controlled_comparison}\subref{fig:efficiency_subfig}. Measurements are reported on the LoCoMo test set (8 conversations, ${\sim}$5{,}000 turns, ${\sim}$1{,}300 questions). Both methods use the same Qwen2.5-7B model, hybrid retrieval pipeline, and answer prompts; the LLM manager baseline performs autoregressive generation for each turn, whereas \ours performs only frozen forward passes.}
\label{tab:efficiency_appendix}
\end{table}

\paragraph{Memory Budget Sweep Values.}
Table~\ref{tab:budget_sweep_appendix} reports the exact budget-sweep values underlying Figure~\ref{fig:budget_sweep}.

\begin{table}[ht]
\centering
\small
\begin{tabular}{@{}cccc@{}}
\toprule
\textbf{Store\%} & \textbf{\ours F1} & \textbf{Random F1} & \textbf{$\Delta$} \\
\midrule
94.5\% & 53.7 & 52.7 & +1.0 \\
80.4\% & 53.2 & 47.4 & +5.8 \\
64.9\% & 52.0 & 41.8 & +10.2 \\
45.9\% & 46.1 & 34.8 & +11.4 \\
19.4\% & 33.5 & 23.3 & +10.2 \\
\bottomrule
\end{tabular}
\caption{Exact values underlying Figure~\ref{fig:budget_sweep}. Each row reports the matched storage ratio, the resulting F1 (\%) for \ours and random storage, and the absolute F1-point gap between the two policies.}
\label{tab:budget_sweep_appendix}
\end{table}

\section{Additional Ablations}
\label{app:additional_results}
This section collects supporting ablations that complement, but are not required to read, the main-paper results.

\paragraph{Retrieval and Prompt Components.}
Table~\ref{tab:retrieval_ablation} reports the incremental ablation over retrieval and prompting components. All configurations use the same trained router; only the retrieval and prompting pipeline changes.

\begin{table}[ht]
\centering
\small
\begin{tabular}{@{}lcccc@{}}
\toprule
\textbf{Configuration} & \textbf{Single} & \textbf{Multi} & \textbf{Temp.} & \textbf{Open} \\
\midrule
Dense only (MiniLM-L6-v2) & 29.5 & 18.9 & 11.0 & 13.9 \\
\ + BGE-large embedding & 44.1 & 33.8 & 19.4 & 12.8 \\
\ + BM25 hybrid retrieval & 44.5 & 35.4 & 19.9 & 14.2 \\
\ + Speaker/temporal boost & 44.5 & 35.4 & 19.9 & 14.2 \\
\ + Speaker-grouped prompt & 54.8 & 26.1 & 34.5 & 17.8 \\
\ + Category-specific prompts & \textbf{57.5} & \textbf{52.4} & \textbf{44.0} & \textbf{27.1} \\
\bottomrule
\end{tabular}
\caption{Incremental ablation of retrieval and prompt components. Starting from dense-only retrieval with MiniLM-L6-v2, each row adds one retrieval or prompting component while keeping the trained router fixed. All evaluations use Qwen2.5-7B as the answer agent, and F1 excludes adversarial questions.}
\label{tab:retrieval_ablation}
\end{table}

The two most impactful changes are: (1) upgrading the retrieval embedding from MiniLM-L6-v2 (22M parameters, 384 dimensions) to BGE-large-en-v1.5~\citep{xiao2023cpack} (335M parameters, 1024 dimensions), which improves single-hop F1 by 14.6 points; and (2) introducing category-specific prompts, particularly the multi-hop prompt that removes the ``5--6 words'' constraint and instructs list-style answers, which lifts multi-hop F1 from 26.1 to 52.4. The speaker-grouped memory presentation substantially helps temporal questions (19.9 $\to$ 34.5) by organizing memories chronologically within each speaker's history.

\paragraph{Factor Attribution.}
Table~\ref{tab:factorial} reports descriptive factor attribution via factorial averaging over admission policy, retrieval method, and prompt style.

\begin{table}[t]
\centering
\small
\begin{tabular}{@{}llcc@{}}
\toprule
\textbf{Factor} & \textbf{Level} & \textbf{Mean F1} & \textbf{$\Delta$ vs.\ baseline} \\
\midrule
\multirow{5}{*}{Admission policy} & Random & 38.7 & ---  \\
 & MLP-only & 47.9 & +9.2 \\
 & Keyword heuristic & 48.3 & +9.6 \\
 & \textbf{\ours} & \textbf{49.0} & \textbf{+10.3} \\
\cmidrule(lr){2-4}
 & \emph{Store-all}$^\dagger$ & 50.5 & +11.8 \\
\midrule
\multirow{2}{*}{Prompt style} & Generic & 44.3 & ---  \\
 & \textbf{Category-specific} & \textbf{49.5} & \textbf{+5.2} \\
\midrule
\multirow{2}{*}{Retrieval} & Cosine-only & 46.5 & ---  \\
 & \textbf{Hybrid (dense+BM25)} & \textbf{47.2} & \textbf{+0.7} \\
\bottomrule
\multicolumn{4}{@{}l@{}}{\footnotesize $\Delta$ is measured relative to the first row in each factor block.} \\
\multicolumn{4}{@{}l@{}}{\footnotesize $^\dagger$ Store-all is shown separately because it is not budget-matched.} \\
\end{tabular}
\caption{Descriptive factor attribution via factorial averaging over 20 configurations ($5$ admission policies $\times$ $2$ retrieval variants $\times$ $2$ prompt styles). Each row reports the mean F1 for one factor level after averaging over all settings of the other two factors. All admission policies are budget-matched to approximately $62\%$ storage except store-all.}
\label{tab:factorial}
\end{table}

\textbf{Admission policy} shows the largest budget-matched contrast in this descriptive averaging: \ours reaches mean F1 49.0, compared with 38.7 for random storage ($+10.3$), while unconstrained store-all reaches 50.5 overall. \textbf{Prompt style} shows the next-largest contrast, with category-specific prompting improving mean F1 by $+5.2$ over the generic prompt, reflecting the importance of answer-format alignment with the evaluation metric. \textbf{Retrieval method} shows the smallest contrast ($+0.7$), indicating that our hybrid retrieval provides only a modest improvement over cosine-only search in this setting. These descriptive averages suggest that write-side admission and prompt design matter more than the retrieval variant in our setup.

\paragraph{Router Training Approaches.}
Table~\ref{tab:training_ablation} compares our supervised training approach against RL-based alternatives. GRPO without supervised warm-starting catastrophically fails (F1 = 1.1): the randomly initialized router overwhelmingly selects NOOP, resulting in empty memory stores and zero QA reward, providing no learning signal. Even with supervised warm-starting followed by GRPO, the RL phase does not improve over supervised-only training (28.5 vs.\ 52.0), as the sparse binary EM reward is insufficient to refine the router's already-reasonable policy. We also attempted training with UPDATE labels (marking semantically similar consecutive turns as updates to existing entries), but this caused the router to over-predict UPDATE at the expense of ADD, collapsing the memory store.

\begin{table}[ht]
\centering
\small
\begin{tabular}{@{}lcc@{}}
\toprule
\textbf{Training Approach} & \textbf{F1 (excl.\ adv.)} & \textbf{Failure Mode} \\
\midrule
GRPO only & 1.1 & Cold-start: all NOOP \\
Supervised + GRPO & 28.5 & GRPO did not improve \\
Supervised (ADD/UPDATE/NOOP) & 1.4--12.8 & UPDATE over-predicted \\
Supervised (ADD/NOOP) & 52.0 & ---  \\
\bottomrule
\end{tabular}
\caption{Router training approach comparison. All variants are evaluated with the same retrieval and prompt pipeline; only the router-training procedure changes. This table highlights the cold-start failure of pure RL and the mode-collapse behavior observed when UPDATE labels are introduced. F1 excludes adversarial questions.}
\label{tab:training_ablation}
\end{table}

\paragraph{Backbone Scaling.}
An advantage of \ours's decoupled architecture is that the answer agent can be swapped independently of the router. Table~\ref{tab:scaling} compares Qwen2.5-7B and Qwen3.5-35B-A3B (4-bit) as answer agents, both using identical router weights. The larger model improves all categories except open-domain: single-hop rises from 57.5 to 60.9, multi-hop from 52.4 to 56.1, and temporal from 44.0 to 49.7. The open-domain decline (27.1 $\to$ 23.8) may be attributable to 4-bit quantization effects on inference-heavy questions.

\begin{table}[ht]
\centering
\small
\begin{tabular}{@{}lccccc@{}}
\toprule
\textbf{Answer Agent} & \textbf{F1 (all)} & \textbf{Single} & \textbf{Multi} & \textbf{Temp.} & \textbf{Open} \\
\midrule
Qwen2.5-7B & 52.0 & 57.5 & 52.4 & 44.0 & \textbf{27.1} \\
Qwen3.5-35B (4-bit) & \textbf{55.5} & \textbf{60.9} & \textbf{56.1} & \textbf{49.7} & 23.8 \\
\bottomrule
\end{tabular}
\caption{Backbone scaling ablation. The router is trained once with Qwen2.5-7B and then kept frozen while only the answer agent changes. This isolates how much downstream answer quality improves when the same write-side policy is paired with a larger LLM. F1 excludes adversarial questions.}
\label{tab:scaling}
\end{table}

\section{Limitations and Future Work}
Our current system has three main limitations. First, open-domain performance remains an area for improvement, suggesting that the frozen answer agent is weaker at multi-step inference from indirect evidence. Second, the router currently learns only ADD and NOOP: attempts to incorporate UPDATE led to mode collapse, highlighting the difficulty of embedding-level memory consolidation. Third, supervised training depends on teacher-generated labels, which introduces an upfront labeling cost; reducing this dependency through weaker supervision or self-supervised objectives is an important direction for future work.

\end{document}